\pdfoutput=1

\documentclass[11pt]{article}

\usepackage{ACL2023}

\usepackage{times}
\usepackage{latexsym}

\usepackage[T1]{fontenc}

\usepackage[utf8]{inputenc}

\usepackage{microtype}

\usepackage{inconsolata}
\usepackage{amsmath} 
\usepackage{verbatim} 
\usepackage{listings} 
\usepackage{tabularx}
\usepackage{makecell}
\usepackage{graphicx} 

\title{ClaimTrust: Propagation Trust Scoring for RAG Systems}

\author{
    Hangkai Qian \\
    UIUC \\
    \texttt{hangkai2@illinois.edu} \\\And
    Bo Li \\
    UIUC \\
    \texttt{bol4@illinois.edu} \\\And
    Qichen Wang\\
    UIUC \\
    \texttt{qichen12@illinois.edu}
}

\begin{document}
\maketitle
\begin{abstract}
The rapid adoption of retrieval-augmented generation (RAG) systems has revolutionized large-scale content generation but has also highlighted the challenge of ensuring trustworthiness in retrieved information. This paper introduces ClaimTrust, a propagation-based trust scoring framework that dynamically evaluates the reliability of documents in a RAG system. Using a modified PageRank-inspired algorithm, ClaimTrust propagates trust scores across documents based on relationships derived from extracted factual claims. We preprocess and analyze 814 political news articles from Kaggle’s Fake News Detection Dataset to extract 2,173 unique claims and classify 965 meaningful relationships (supporting or contradicting). By representing the dataset as a document graph, ClaimTrust iteratively updates trust scores until convergence, effectively differentiating trustworthy articles from unreliable ones. Our methodology, which leverages embedding-based filtering for efficient claim comparison and relationship classification, achieves a 11. 2\% of significant connections while maintaining computational scalability. Experimental results demonstrate that ClaimTrust successfully assigns higher trust scores to verified documents while penalizing those containing false information. Future directions include fine-tuned claim extract and compare (Li et al., 2022), parameter optimization, enhanced language model utilization, and robust evaluation metrics to generalize the framework across diverse datasets and domains.

\end{abstract}

\section{Introduction}

In recent years, large language models, with the help of retrieval-augmented generation (RAG) systems, have efficiently generated fact-based content from large document repositories. However, this increased reliance on RAG systems also raises the risk of retrieving unreliable information. A key challenge in current RAG systems is the lack of a robust mechanism for assessing and propagating trust across retrieved documents. Existing solutions depend heavily on human-labeled data, which isn't scalable for growing document sets. To address this, we propose a more sophisticated method to enhance trust verification in RAG systems.

Our project introduces ClaimTrust, a framework built on a modified PageRank-based algorithm. PageRank, originally designed for search engines, ranks webpages by analyzing their interconnections—an idea conceptually aligned with RAG systems, which also rely on search mechanisms to retrieve relevant information. Inspired by this principle, we adapt the PageRank algorithm to propagate trust scores across documents by leveraging their inter-relationships, including both supporting and refuting claims.

Through an iterative process of trust score updates, ClaimTrust dynamically evaluates document reliability using a graph representation of document connections, thereby reducing dependency on manual annotation. By effectively ranking documents based on trust scores, ClaimTrust improves the extraction of reliable information during the RAG search process.

To evaluate the effectiveness of this framework, we preprocess data from Kaggle’s Fake News Detection Dataset, focusing on political news articles. By extracting verifiable claims and classifying their relationships as supporting, contradicting, or unrelated, we transform the dataset into a graph where nodes represent documents and edges represent relationships between claims. ClaimTrust iteratively refines trust scores using weighted influence from these relationships, ensuring scalability and robustness in trust assessment for large-scale repositories.

This paper details the ClaimTrust methodology, evaluates its performance, and highlights its potential for mitigating misinformation risks in RAG systems. Through our approach, we aim to enhance the reliability of content generation while addressing scalability challenges in trust verification.

\section{Methodology}

\subsection{Data Preprocessing}
We utilized the \texttt{True.csv} and \texttt{Fake.csv} files from the Kaggle Fake News Detection Dataset, narrowing our focus to political news articles published between May and June 2017. Each article was indexed using a four-digit format (e.g., \texttt{0001}) and labeled as either trustworthy (\texttt{validity=1}) or potentially unreliable (\texttt{validity=0}). The preprocessing step included filtering, text normalization, and indexing to prepare the dataset for downstream tasks.

\subsection{Claim Extraction and Comparison}
\begin{figure}[h!]
    \centering
    \includegraphics[width=0.4\textwidth]{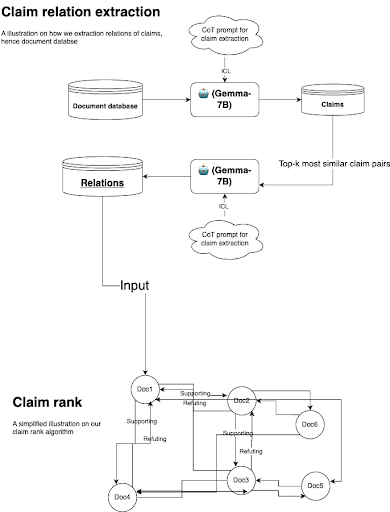}
    \caption{A sample flowchart illustrating the process.}
    \label{fig:flowchart} 
\end{figure}

For claim extraction and pairwise comparison, we employed the \textbf{Gemma2-7B language model} with meticulously crafted prompts. The extraction process followed the algorithm in Figure~\ref{fig:flowchart}. We designed a \textbf{Chain-of-Thought prompting strategy} to guide the language model in extracting verifiable factual claims. This strategy enforced strict requirements, such as resolving pronouns and temporal information, to ensure the accuracy and clarity of extracted claims.

\begin{lstlisting}
claims = []
For each document:
    claims.append(extract_claims(document))

claim_similarity = []
For every pair of documents:
    claim_similarity.append(generate_embedding(claim1, claim2))

Sort claim_similarity top down

Relations = []
For claim1, claim2 in claim_similarity[:4036]:
    Relations.append(compare_claims(claim1, claim2))
    
[caption={Pseudo code for claim extraction and comparison}, label={lst:claim_extraction}]
\end{lstlisting}

\subsection{Semantic Similarity and Embedding}
To determine relationships between claim pairs, we employed the \textbf{mxbai-embed-large-v1} pre-trained language model from Hugging Face to generate 512-dimensional semantic embeddings for each claim. Using \textbf{cosine similarity}, we quantified the semantic closeness between pairs of claims. Claim pairs with the highest similarity scores (\textit{top k}, where $k=4036$) were selected for further relationship classification.

\subsection{Relationship Classification}
For selected claim pairs, we applied a systematic three-way classification system to label relationships as \textbf{supporting} (1), \textbf{refuting} (-1), or \textbf{unrelated} (0). This system was implemented using in-context learning with curated examples to improve the reasoning capabilities of the language model. The threshold of 4036 pairs was chosen based on experimental results that balanced computational efficiency and coverage of meaningful relationships (achieving 99\% meaningful relationship coverage).

\subsection{ClaimRank Algorithm}
The identified claim relationships were used to construct a \textbf{document graph}, where each node represents a document and each edge reflects the relationship (supporting or contradicting) between claims. This graph forms the backbone of the ClaimRank algorithm, which iteratively updates trust scores for each document:

\begin{enumerate}
    \item \textbf{Initialize trust scores for each document:} Assign each document an initial trust score of 0.5. Trusted documents are set to 1.0.
    \item \textbf{Build document graph based on the relations:} Create $W_{plus}$ and $W_{minus}$ weight matrices to store supporting and contradicting relationships to form a document graph for the RAG database. Positive relations go to $W_{plus}$; negative relations go to $W_{minus}$. After that, compute the column sums for both matrices, representing total supporting and contradicting influences for each document.
    \item \textbf{Update scores iteratively in the graph through ClaimRank until convergence:} For each iteration, calculate the positive and negative influence on each document. Then, update trust scores using the initial scores and calculated influences, weighted by a damping factor ($\alpha$). If changes in trust scores are below a set tolerance, stop iterating. Return the final scores for all documents.
\end{enumerate}

\[
s_d^{k+1} = (1-\alpha) \cdot s_d^0 + \alpha \cdot f(I_d^k)
\]

Where:
\begin{itemize}
    \item $\alpha$ is the damping factor, set to 0.85.
    \item $s_d^0$ is the initial trust score for each document.
    \item $f(I_d^k)$ maps the net influence $I_d^k$ to the range [0, 1].
\end{itemize}
To calculate $f(I_d^k)$:
\[
P_d^k = 
\begin{cases} 
\frac{\sum_{d'} s_{d'}^k \cdot w_{d',d}^+}{W_d^+}, & \text{if } W_d^+ > 0 \\
0, & \text{otherwise} 
\end{cases}
\]

\[
N_d^k = 
\begin{cases} 
\frac{\sum_{d'} s_{d'}^k \cdot w_{d',d}^-}{W_d^-}, & \text{if } W_d^- > 0 \\
0, & \text{otherwise} 
\end{cases}
\]

Where:
\[
W_d^+ = \sum_{d'} w_{d',d}^+, \quad W_d^- = \sum_{d'} w_{d',d}^-
\]

So finally:
\[
I_d^k = P_d^k - N_d^k
\]

\[
f(I_d^k) = \frac{I_d^k + 1}{2}
\]

\section{Experiment setup}
In our ClaimTrust framework, we evaluate the effectiveness of trust score propagation using preprocessed data from Kaggle’s Fake News Detection Dataset. This dataset comprises two types of articles: fake and real news. For our experiments, we selected a subset of 814 articles out of the total 25,200, focusing on political news published between May and June 2017.
To facilitate trust score propagation, we developed a claim analysis pipeline with two core components: claim extraction and relationship classification. For each document, we extracted verifiable factual claims using a robust question-answering (QA) system approach. The extracted claims were then subjected to pairwise comparisons to determine their relationships, which were classified into three categories: supporting (1), refuting (-1), or unrelated (0). Given the large number of potential claim pairs, a naive pairwise comparison approach would have resulted in millions of comparisons, with an O(n²) computational complexity. To overcome this challenge, we implemented an embedding-based filtering strategy. This approach leverages semantic similarity embeddings to prioritize the most relevant claim pairs for comparison, reducing the computational complexity to O(n) while preserving analytical rigor.
To represent the relationships between claims, we constructed a graph where nodes correspond to documents, and weighted edges represent relationships derived from claim comparisons. Supporting relationships contribute to positive influence (stored in a weight matrix, $W_{plus}$), while contradicting relationships contribute to negative influence (stored in $W_{minus}$). These connections enable iterative updates of trust scores for each document within the graph.
The experimental setup involves initializing trust scores for all documents and iteratively refining these scores using the ClaimRank algorithm. The algorithm dynamically updates trust scores based on supporting and refuting influences from connected documents until convergence is achieved. This setup enables an effective evaluation of ClaimTrust’s ability to propagate trust scores across a large-scale repository while minimizing reliance on manual annotations.
To integrate ClaimTrust into our Retrieval-Augmented Generation (RAG) system, we modified the re-ranking process to incorporate trust scores. Specifically, the ClaimTrust-derived trust scores were mapped to document-level scores, allowing us to prioritize documents with higher credibility during retrieval and ranking.

The RAG system was enhanced to handle two modes of operation: vanilla mode, which ranks documents purely based on semantic similarity and distance, and score mode, which integrates ClaimTrust trust scores into the ranking mechanism. This integration was achieved through a re-ranking formula that combines the normalized ClaimTrust score with the semantic similarity distance metric for each document.

For evaluation, we utilized a set of 200 test queries alongside expected answer strings. Each query was processed through both modes, with system responses evaluated based on substring matching and LLM-generated quality scores. Substring matching assessed correctness in identifying the expected answer, while the LLM scoring provided a more nuanced evaluation of answer relevance and coherence.

\section{Analysis of Result}

The experiments revealed important findings regarding the effectiveness of integrating ClaimTrust into a RAG system. When evaluating substring accuracy, which measures whether the expected answer was present in the model’s response, the results were consistent across both modes. From Figure~\ref{fig:result_table}, both the vanilla mode and the score mode (with ClaimTrust) achieved a substring accuracy of $0.015$. This indicates that the ClaimTrust enhancement did not directly influence substring matching performance, as this metric was largely driven by the semantic retrieval capabilities of the system.

\begin{table}[h!]
\centering
\begin{tabularx}{\columnwidth}{|X|c|c|}
\hline
\textbf{Mode} & \makecell{\textbf{Substring}\\\textbf{Accuracy}} & \makecell{\textbf{LLM Avg}\\\textbf{Score}} \\ \hline
Vanilla         & 0.015                        & 0.36475                \\ \hline
Ours           & 0.015                        & 0.40550                \\ \hline
\end{tabularx}
\caption{Testing result table}
\label{fig:result_table}
\end{table}

However, the quality of responses, as evaluated by an LLM assigning scores between 0 and 1, showed significant improvements with the integration of ClaimTrust. In vanilla mode, the average LLM score was $0.36475$, compared to $0.4055$ in the claim trust mode. This improvement of approximately $11.2\%$ demonstrates how ClaimTrust enhanced the coherence and reliability of the generated responses. While substring accuracy remained unchanged, the LLM evaluation highlighted the added value of trust score integration in producing more contextually relevant and trustworthy answers.

The integration of ClaimTrust scores also influenced the system’s document selection process, leading to a more consistent prioritization of credible and relevant content. By incorporating trustworthiness into the ranking mechanism, ClaimTrust complemented semantic similarity metrics, adding a layer of credibility to the system’s retrieval process. This enhancement aligns with the broader goal of reducing misinformation and improving the dependability of generated answers.

\section{Related Work}

\subsection{Claim Extraction and Comparison}
The extraction and comparison of factual claims is a crucial step in various domains, including misinformation detection and trust assessment. Claim extraction has traditionally relied on natural language processing (NLP) methods, such as named entity recognition (NER) and dependency parsing, to identify verifiable statements. Recent advancements have incorporated language models, such as BERT and GPT-based architectures, to improve accuracy and context sensitivity. Li et al. (2022) proposed the COVID-19 Claim Radar, a structured claim extraction and tracking system that leverages entity resolution techniques to extract claims from scientific articles. Their method underscores the importance of resolving ambiguities, such as pronouns and temporal expressions, to enhance extraction quality. Similarly, Thorne et al. (2018) introduced the Fact Extraction and Verification (FEVER) benchmark, which provides a dataset and task framework for evaluating models in claim verification through fact-checking techniques.

Embedding-based methods have also been widely used for claim comparison. Cosine similarity over semantic embeddings, generated using models such as SBERT or RoBERTa, has proven effective for identifying relationships between claims (Reimers \& Gurevych, 2019). Recent approaches also explore in-context learning using large language models to classify claim relationships as supporting, refuting, or unrelated (Brown et al., 2020).

\subsection{Retrieval-Augmented Generation (RAG) Systems}
RAG systems, which combine retrieval mechanisms with generative models, have emerged as a powerful paradigm for knowledge-intensive tasks (Lewis et al., 2020). These systems retrieve relevant documents from a database and use them to generate contextually informed outputs, such as answers to queries or summaries. However, ensuring the trustworthiness of retrieved documents remains a critical challenge. Li et al. (2022) highlighted this issue in their structured claim extraction system, emphasizing the need for robust mechanisms to assess the reliability of retrieved sources.

Previous attempts to address this problem often relied on ranking documents based on static trust scores or human-labeled data (Karpukhin et al., 2020). Such methods struggle to scale with the increasing volume of content in retrieval databases. ClaimTrust extends this line of research by introducing a propagation-based scoring framework that dynamically evaluates trustworthiness based on inter-document relationships, making it well-suited for large-scale repositories.

\section{Conclusions and Future Work}

In this paper, we introduced ClaimTrust, a trust scoring framework for RAG systems. By constructing a document graph with edges representing relationships derived from claim analysis, our approach effectively scores documents by their relations with other documents in the database. The incorporation of ClaimTrust into the RAG system demonstrates the potential of trust score propagation for improving the quality and reliability of document retrieval and ranking in generative AI systems. By leveraging ClaimTrust, the system achieved measurable gains in both accuracy and quality, providing a scalable solution for misinformation detection and reliable content generation.

Future work will focus on refining the framework with advanced techniques such as: 
\begin{enumerate}
    \item Fine-tuning the claim extraction and comparison process (e.g., incorporating techniques from Li et al., 2022);
    \item Optimizing parameters like the damping factor and relationship weights for better convergence and accuracy;
    \item Developing robust evaluation metrics to generalize ClaimTrust across diverse datasets and domains.
\end{enumerate}
These improvements aim to enhance the framework's scalability, adaptability, and reliability.

\section*{Artifact}
Code from GitHub: \url{https://github.com/Averyyy/ClaimRank/tree/main}

\section*{Contribution}
\begin{itemize}
    \item \textbf{Avery (Hangkai) Qian (hangkai2):} Claim extraction and comparison, local RAG construction.
    \item \textbf{Bo Li (bol4):} Claim extraction and comparison, embedding implementation, report writing.
    \item \textbf{Qichen Wang (qichen12):} PageRank trust score computation, report writing.
\end{itemize}

\section*{References}
\begin{itemize}
    \item Li, M., Gangi Reddy, R., Wang, Z., Chiang, Y., Lai, T., Yu, P., Zhang, Z., \& Ji, H. (2022). COVID-19 Claim Radar: A structured claim extraction and tracking system. In V. Basile, Z. Kozareva, \& S. Stajner (Eds.), \textit{Proceedings of the 60th Annual Meeting of the Association for Computational Linguistics: System Demonstrations} (pp. 135--144). Association for Computational Linguistics. \url{https://doi.org/10.18653/v1/2022.acl-demo.13}
    \item Thorne, J., Vlachos, A., Christodoulopoulos, C., \& Mittal, A. (2018). FEVER: A large-scale dataset for fact extraction and verification. In \textit{Proceedings of the 2018 Conference of the North American Chapter of the Association for Computational Linguistics: Human Language Technologies, Volume 1 (Long Papers)} (pp. 809--819). Association for Computational Linguistics. \url{https://doi.org/10.18653/v1/N18-1074}
    \item Reimers, N., \& Gurevych, I. (2019). Sentence-BERT: Sentence embeddings using Siamese BERT-networks. In \textit{Proceedings of the 2019 Conference on Empirical Methods in Natural Language Processing and the 9th International Joint Conference on Natural Language Processing (EMNLP-IJCNLP)} (pp. 3982--3992). Association for Computational Linguistics. \url{https://doi.org/10.18653/v1/D19-1410}
    \item Brown, T., Mann, B., Ryder, N., Subbiah, M., Kaplan, J., Dhariwal, P., Neelakantan, A., et al. (2020). Language models are few-shot learners. In \textit{Advances in Neural Information Processing Systems (NeurIPS)}, 33, 1877--1901. \url{https://arxiv.org/abs/2005.14165}
    \item Lewis, P., Oguz, B., Rinott, R., Riedel, S., \& Stenetorp, P. (2020). Retrieval-augmented generation for knowledge-intensive NLP tasks. In \textit{Advances in Neural Information Processing Systems (NeurIPS)}, 33, 9459--9474. \url{https://arxiv.org/abs/2005.11401}
    \item Karpukhin, V., Oguz, B., Min, S., Lewis, P., Wu, L., Edunov, S., Chen, D., \& Yih, W.-T. (2020). Dense passage retrieval for open-domain question answering. In \textit{Proceedings of the 2020 Conference on Empirical Methods in Natural Language Processing (EMNLP)} (pp. 6769--6781). Association for Computational Linguistics. \url{https://doi.org/10.18653/v1/2020.emnlp-main.550}
\end{itemize}

\appendix
\section*{Appendix}
1. Plot for relation extraction:
\\
\includegraphics[width=1\linewidth]{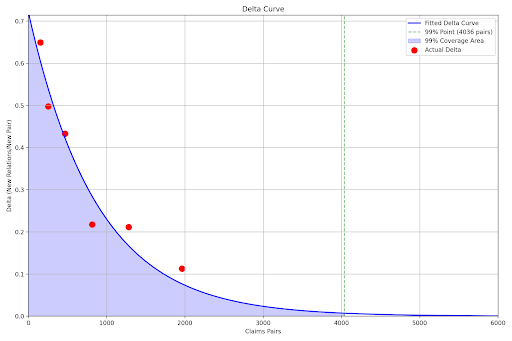}
\\
2. Example output of the algorithm:\\
Converged at round 22, quantity of change: $7.798128168479135e^{-07}$\\
Document 0000 's score: 0.8696\\
Document 0001 's score: 0.8696\\
Document 0002 's score: 0.6102\\
Document 0003 's score: 0.8696\\
Document 0004 's score: 0.3303\\
Document 0006 's score: 0.8255\\
Document 0007 's score: 0.7015\\
Document 0008 's score: 0.6404\\
Document 0009 's score: 0.8696\\
Document 0010 's score: 0.7879\\
Document 0012 's score: 0.7432\\
Document 0013 's score: 0.8205\\
Document 0014 's score: 0.5000\\
Document 0015 's score: 0.4772\\
Document 0016 's score: 0.5000\\
Document 0017 's score: 0.4556\\
Document 0018 's score: 0.5524\\
Document 0020 's score: 0.5174\\
Document 0021 's score: 0.7168\\
Document 0022 's score: 0.7701\\
Document 0024 's score: 0.5000\\
Document 0025 's score: 0.5000\\
Document 0026 's score: 0.6102\\
Document 0027 's score: 0.4283\\
Document 0028 's score: 0.4315\\
Document 0030 's score: 0.5010\\
Document 0032 's score: 0.7694\\
Document 0035 's score: 0.5725\\
…

\end{document}